\documentclass{article}

\usepackage[final]{nips_2016}

\usepackage[utf8]{inputenc} 
\usepackage[T1]{fontenc}    
\usepackage[hidelinks]{hyperref}       
\usepackage{url}            
\usepackage{booktabs}       
\usepackage{amsfonts}       
\usepackage{nicefrac}       
\usepackage{microtype}      

\usepackage{pgfplots,tikz}
    \usepgfplotslibrary{dateplot}
\usepackage[USenglish]{babel}
\usepackage[nolist,nohyperlinks]{acronym}
\usepackage{amsmath,amssymb,mathtools,dsfont,physics}
\usepackage{siunitx}

\newcommand{\R}{\mathds{R}}
\newcommand{\N}{\mathds{N}}
\DeclareMathOperator{\tril}{tril}
\DeclareMathOperator{\diag}{diag}

\acrodef{sph}[SPH]{Smoothed-Particle Hydrodynamics}
\acrodef{cnn}[CNN]{Convolutional Neural Network}
\acrodef{ppe}[PPE]{Pressure Poisson Equation}
\acrodef{spd}[SPD]{Symmetric Positive Definite}
\acrodef{cg}[CG]{Conjugate Gradient}
\acrodef{pcg}[PCG]{Preconditioned Conjugate Gradient}
\acrodef{svd}[SVD]{Singular Value Decomposition}
\acrodef{cfd}[CFD]{Computational Fluid Dynamics}
\acrodef{ml}[ML]{Machine Learning}
\acrodef{gpu}[GPU]{Graphics Processing Unit}
\acrodef{pde}[PDE]{Partial Differential Equation}
\acrodef{prelu}[PReLU]{Parametric Rectified Linear Unit}
\acrodef{amg}[AMG]{Algebraic MultiGrid}

\title{Deep Learning of Preconditioners for Conjugate Gradient Solvers in Urban Water Related Problems}

\author{
    Johannes~Sappl\thanks{corresponding author}\\
    Unit of Environmental Engineering\\
    Faculty of Engineering Sciences\\
    Universit\"at Innsbruck\\
    Technikerstra\ss e~13, 6020 Innsbruck\\
    Austria\\
    \href{mailto:johannes.sappl@uibk.ac.at}{johannes.sappl@uibk.ac.at}\\
    \And
    Laurent~Seiler\\
    Unit of Environmental Engineering\\
    Faculty of Engineering Sciences\\
    Universit\"at Innsbruck\\
    Technikerstra\ss e~13, 6020 Innsbruck\\
    Austria\\
    \href{mailto:laurent.seiler@uibk.ac.at}{laurent.seiler@uibk.ac.at},
    \And
    Matthias~Harders\\
    Interactive Graphics and Simulation Group\\
    Department of Computer Science\\
    Universit\"at Innsbruck\\
    Technikerstra\ss e~21~A, 6020 Innsbruck\\
    Austria\\
    \href{mailto:matthias.harders@uibk.ac.at}{matthias.harders@uibk.ac.at}
    \And
    Wolfgang~Rauch\\
    Unit of Environmental Engineering\\
    Faculty of Engineering Sciences\\
    Universit\"at Innsbruck\\
    Technikerstra\ss e~13, 6020 Innsbruck\\
    Austria\\
    \href{mailto:wolfgang.rauch@uibk.ac.at}{wolfgang.rauch@uibk.ac.at},
}

\begin{document}

\maketitle

\begin{abstract}
    Solving systems of linear equations is a problem occuring frequently in water engineering applications. Usually the size of the problem is too large to be solved via direct factorization. One can resort to iterative approaches, in particular the conjugate gradients method if the matrix is symmetric positive definite. Preconditioners further enhance the rate of convergence but hitherto only handcrafted ones requiring expert knowledge have been used. We propose an innovative approach employing Machine Learning, in particular a Convolutional Neural Network, to unassistedly design preconditioning matrices specifically for the problem at hand. Based on an in-depth case study in fluid simulation we are able to show that our learned preconditioner is able to improve the convergence rate even beyond well established methods like incomplete Cholesky factorization or Algebraic MultiGrid.
\end{abstract}

\section{Introduction}

Given $\vb{A} \in \R^{m\times n}$, $\vb{b}\in\R^m$ for some positive $m$, $n \in \N$ a system of $m$ linear equations with $n$ variables can be written as
\begin{equation}\label{eq:linSystem}
    \vb{A} \vb{x} = \vb{b}\,,
\end{equation}
where $\vb{x} \in\R^n$ denotes some yet unknown solution. This type of problem frequently appears in urban water engineering and management applications, e.\,g., when solving an optimization problem using the method of least squares. We are particularly interested in linear systems~\eqref{eq:linSystem} where $\vb{A} \in \R^{n\times n}$ is \ac{spd} as well as sparse. Examples are finite-difference methods for space-time discretization of a \ac{pde} in storm water modeling or computing flow continuity and head loss equations characterizing the hydraulic state of a pipe network as, e.\,g., in the software EPANET~2~\citep{rossman2000epanet}. 
Very often $n$ can be quite large with millions or even billions of unknowns to solve for. In this case direct methods like Cholesky decomposition become infeasible since they scale poorly with the size of the underlying problem in terms of operations and memory needed~\citep{benzi2002}.

Iterative approaches which improve approximate solutions based on previous estimates provide a remedy for this, albeit at the expense of accuracy. One of the best known iterative methods for solving sparse \ac{spd} systems is the \ac{cg} algorithm to which the following well-known a priori error bound applies. Let $\vb{x}_j$ be the approximate solution obtained after the $j$-th \ac{cg}-iteration, then the norm of the error depending on $\vb{A}$ is limited by
\begin{equation}\label{eq:errorBound}
    \norm{\vb{x} - \vb{x}_j}_{\vb{A}} \leq 2 \Bigg[ \frac{\sqrt{\kappa(\vb{A})} -1}{\sqrt{\kappa(\vb{A})} + 1} \Bigg]^j \norm{ \vb{x} - \vb{x}_0}_{\vb{A}}\,,
\end{equation}
for some initial guess $\vb{x}_0$~\citep{saad2003iterative}, where the so-called condition number of $\vb{A}$ is defined as
\begin{equation}\label{eq:kappaDef}
    \kappa(\vb{A}) \coloneqq \frac{ \sigma_{\max}(\vb{A}) }{ \sigma_{\min}(\vb{A}) } \geq 1\,,
\end{equation}
and $\sigma_{\max}(\vb{A})$, $\sigma_{\min}(\vb{A})$ denote the maximum and minimum singular value of $\vb{A}$, respectively. If $\kappa(\vb{A}) \gg 1$ then problem~\eqref{eq:linSystem} is said to be ill-conditioned since, as stated in~\eqref{eq:errorBound}, the asymptotic behavior of $\norm{\vb{x} - \vb{x}_j}_{\vb{A}}$ is determined by $(1+\varepsilon)^{-j}$ for some small $\varepsilon>0$ as $j$ approaches infinity. Based on~\eqref{eq:errorBound} the rate of convergence of \ac{cg} can be improved through preconditioning, e.\,g., from the right, via transforming~\eqref{eq:linSystem} into another linear system with more favorable properties by means of some non-singular \ac{spd} matrix $\vb{M}$ according to
\begin{equation}\label{eq:leftPrecond}
    \vb{A} \vb{M}^{-1} \vb{y} = \vb{b}\,,\quad \vb{x} = \vb{M}^{-1} \vb{y}\,,
\end{equation}
since the solution $\vb{x}$ to~\eqref{eq:linSystem} coincides with the one to~\eqref{eq:leftPrecond}. The so-called preconditioner $\vb{M}$ is chosen s.\,t. it is cheap to construct, apply, and furthermore $1 \leq \kappa(\vb{A} \vb{M}^{-1}) \ll \kappa(\vb{A})$, ultimately resulting in faster convergence of the \ac{pcg} method in agreement with~\eqref{eq:errorBound}. Note that one can also precondition~\eqref{eq:linSystem} from the left $\vb{M}^{-1} \vb{A}$ or even perform split preconditioning $\vb{M}_1^{-1} \vb{A} \vb{M}_2^{-1}$, but since the eigenvalues of the respective matrices are equal, the rate of convergence of \ac{pcg} is going to stay the same. Thus, without loss of generality we are committing ourselves to right preconditioning purely for implementation reasons in the remainder of this paper.

Good designs for the matrix $\vb{M}$ require problem-specific knowledge and are often very situational meaning small changes to the system~\eqref{eq:linSystem} can render $\vb{M}$ inefficient. Instead of manually designing a new preconditioner each time the underlying problem changes, we propose taking advantage of a novel \ac{ml} approach. Perhaps most related to this is a data-driven multigrid method optimizing restriction and prolongation operators for solving discretized \ac{pde}s~\citep{katrutsa2017deep}.

\section{Methods}

At the heart of \ac{ml} is a data set $(x_j, y_j)_j$ and a model $f$ with internal parameters that have to be adjusted based on the data s.\,t. $f(x_j) \eqqcolon \hat{y}_j$ has small loss $\ell(\hat{y}_j, y_j)$ for as many samples $j$ as possible. If the ground truths $y_j$ are not available the problem is called unsupervised and $f$ is trained only with respect to $\ell(\hat{y}_j)$. An excellent introduction to \ac{ml} in general is~\citep{shalev2014understanding}. Our theoretical framework includes the definition of an \ac{ml} model
\begin{equation}\label{eq:ml_model}
    f\colon \R^{n\times n} \to \R^{n \times n}\colon \vb{A} \mapsto \vb{M}^{-1}
\end{equation}
and the loss function
\begin{equation}\label{eq:loss}
    \ell( \vb{A} ) \coloneqq \kappa\big( \vb{A} f(\vb{A}) \big) = \kappa\big( \vb{A} \vb{M}^{-1} \big)\,.
\end{equation}
The task of finding a preconditioner can now be formulated as an unsupervised learning problem. Given a size $N$ training data set $\mathcal{T} = (\vb{A}_j)_{j=1}^N$ of \ac{spd} matrices the parameters of~\eqref{eq:ml_model} are optimized with regard to~\eqref{eq:loss} s.\,t.
\begin{equation}\label{eq:loss_function}
    \sum_{j=1}^N \ell\big( f(\vb{A}_j) \big) \to \min
\end{equation}
employing a gradient-based optimization algorithm developed by~\citet{kingma2014adam}. So as not to be restricted by the spatial dimensions of input matrices we define the architecture of $f$ as a so-called \ac{cnn} which is a special class of \ac{ml} models focusing on extracting local features in images via convolution operations~\citep{lecun1989}. In principle a \ac{cnn} operates the same way as the famous approach by~\citet{viola2001rapid} used for face detection in digital cameras. The only difference is that features in the \ac{cnn} are trained on the data, rather than having to be constructed by hand in a meaningful way. The \ac{cnn} architecture allows for $f$ to come up with preconditioners for a wider range of \ac{spd} matrices since the model is invariant with respect to their shapes. The \ac{prelu}
\begin{equation*}\label{eq:prelu}
    \sigma\colon \R \to \R\colon x \mapsto \begin{cases}x \quad &\text{if }x>0 \\
        ax \quad &\text{otherwise}
    \end{cases}
\end{equation*}
where $a \in \R$ represents a trainable parameter, is chosen as the non-linear activation function in-between convolutional layers. Its advantage is that, unlike non-negative activations which crop each value to $[0, \infty)$, negative entries that might occur in the preconditioner $\vb{M}^{-1}$ can easily propagate through $f$.

\begin{figure*}[htb]
    \centering
    \begin{tikzpicture}
        \node[inner sep=0pt] (orig) at (0,0)
        {\includegraphics[width=.15\textwidth]{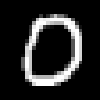}};
        \node[inner sep=0pt] (one) at (3.5,0)
        {\includegraphics[width=.15\textwidth]{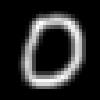}};
        \node[inner sep=0pt] (two) at (7,0)
        {\includegraphics[width=.15\textwidth]{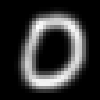}};
        \node[inner sep=0pt] (three) at (10.5,0)
        {\includegraphics[width=.15\textwidth]{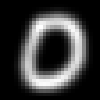}};

        \draw[shorten >=6pt,shorten <=6pt,->,thick] (orig) -- (one) node[midway,above] {$3\times 3$} node[midway,below] {Conv};
        \draw[shorten >=6pt,shorten <=6pt,->,thick] (one) -- (two) node[midway,above] {$3\times 3$} node[midway,below] {Conv};
        \draw[shorten >=6pt,shorten <=6pt,->,thick] (two) -- (three) node[midway,above] {$3\times 3$} node[midway,below] {Conv};

        \node at (0,-1.5) {$\SI{21.56}{\percent}$};
        \node at (3.5,-1.5) {$\SI{37.24}{\percent}$};
        \node at (7,-1.5) {$\SI{51.53}{\percent}$};
        \node at (10.5,-1.5) {$\SI{64.92}{\percent}$};
    \end{tikzpicture}
    \caption{In a left-to-right order each convolution operation (Conv) with a $3\times 3$ kernel further reduces the amount of zeros (black) thus increasing the density (below image) of the matrix. In order to not let this bleeding effect get out of hand our model only has four $2\times 2$ convolutional layers.}\label{fig:bleeding}
\end{figure*}
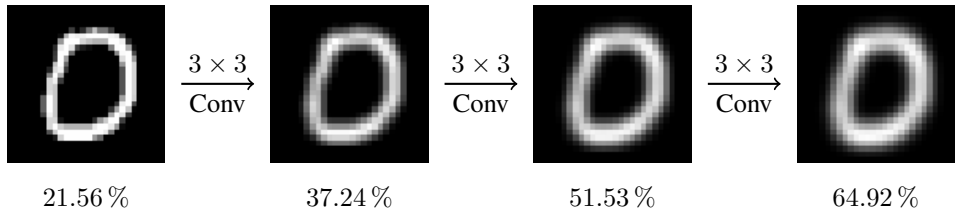

\begin{figure*}[htb]
    \centering
    \includegraphics[clip=true,trim=0pt 65pt 0pt 65pt]{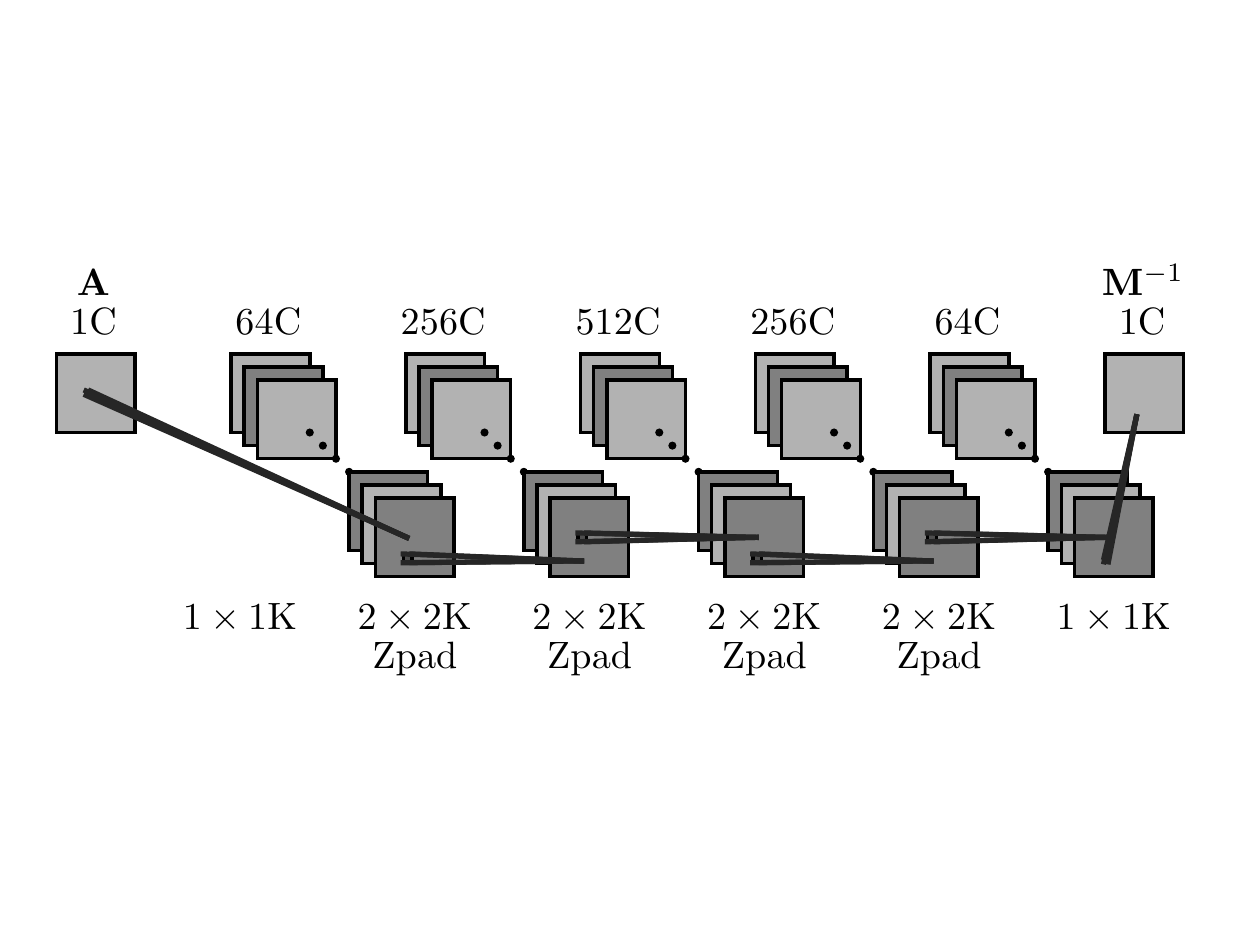}
    \caption{Standard fully convolutional six-layer \ac{cnn} with $1 \times 1$ and $2 \times 2$ convolution kernels (K) stacked in an autoencoder-like structure of channels (C) sweeping over the matrix $\vb{A}$ resulting in a preconditioner $\vb{M}^{-1}$. Due to this architecture $\vb{A}$ can be of arbitrary size. Parameters of this model are optimized according to~\eqref{eq:loss_function}. \ac{prelu} activations are employed in-between layers, zero padding (Zpad) preserves spatial dimensions.}\label{fig:cnn}
\end{figure*}

When evaluating a convolutional layer with window size $k\times k$, $k>1$ for a sparse image, additional non-zero elements are introduced due to the \glqq bleeding\grqq\ effect demonstrated in Figure~\ref{fig:bleeding}.
Hence, the amount of such layers in our \ac{cnn} regulates the density of the learned preconditioner. We introduce merely four $2\times 2$ convolution layers and since $\vb{A}$ is sparse, the sparsity of $\vb{M}^{-1}$ is preserved, resulting in a preconditioner that is cheap to apply. The two $1\times 1$ convolution kernels do generate extra non-zeros. In Figure~\ref{fig:cnn} a schematic representation of the \ac{cnn} model is depicted.

\begin{figure*}[htb]
    \centering
    \includegraphics[width=.6\textwidth]{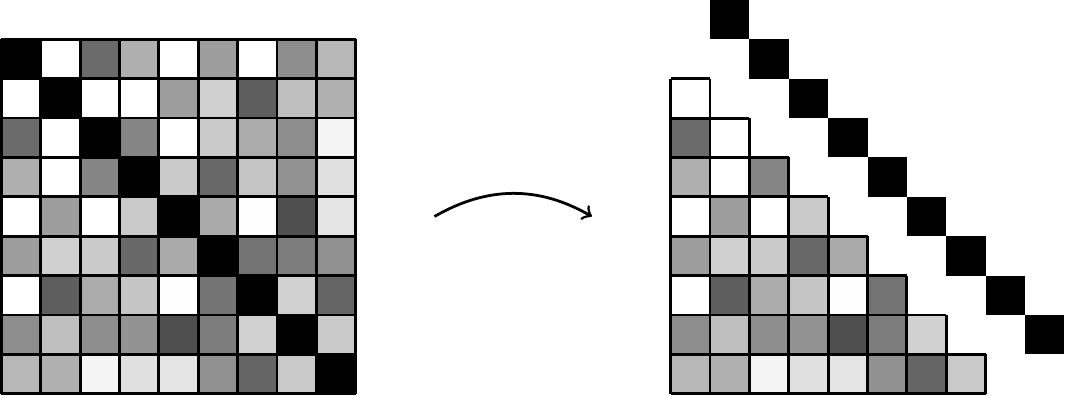}
    \caption{Instead of letting the \ac{cnn} process the entire sparse \ac{spd} matrix $\vb{A}$ (left) which can be quite large, only the strictly lower triangular part $\tril(\vb{A})$ and diagonal $\diag(\vb{A})$ (right) are used as input to reduce the computational burden. A custom \ac{cnn} implementation exploits the sparsity of $\vb{A}$.}\label{fig:input_cnn}
\end{figure*}

The size of images \ac{cnn}s have to process have initially been quite small, e.\,g., $28\times 28$ pixels~\citep{lecun1998gradient}. Only recently have researchers experimented with bigger images like $1024\times 1024$ as well~\citep{karras2017progressive}. Matrices $\vb{A}$ defining linear systems~\eqref{eq:linSystem} exceed such dimensions by far and can cause a \ac{gpu} used for training the model $f$ to be out of memory. Since we are concerned with sparse \ac{spd} matrices, considerable computational power and memory allocated to saving training data $\mathcal{T}$ on the \ac{gpu} would go to waste if sparsity is not properly exploited, resulting among others in evaluation of the convolution kernels on large patches of zeros. Hence, we employ a custom \ac{cnn} implementation based on PyTorch~\citep{paszke2017automatic}, which is freely available on GitHub\footnote{\url{https://github.com/traveller59/spconv}}. Feeding the decomposition of $\vb{A}$ into its strictly lower triangular part $\tril(\vb{A})$ and diagonal $\diag(\vb{A})$ to the \ac{cnn} further reduces the computational burden, as seen in Figure~\ref{fig:input_cnn}.

A Hermitian matrix $\vb{H}$ is positive definite if and only if it has a unique Cholesky decomposition meaning there exists a unique lower triangular matrix $\vb{T}$ with real and strictly positive diagonal elements s.\,t. $\vb{H} = \vb{T}\vb{T}^\ast$. Since the learned preconditioner needs to be \ac{spd} for \ac{pcg} to work, such a unique decomposition has to exist for $\vb{M}^{-1}$ as well. However, the output of the \ac{cnn} model yields the strictly lower triangular part $\vb{T} \coloneqq \tril(f(\vb{A}))$ and the diagonal $\hat{\vb{D}} \coloneqq \diag(f(\vb{A}))$ which does not necessarily have only strictly positive elements. Only after pointwise application of a hard threshold $\vb{D} = \max\{\hat{\vb{D}}, \varepsilon\}$ with $\varepsilon = \num{e-3}$ is
\begin{equation*}
    \vb{M}^{-1} = (\vb{T} + \vb{D})(\vb{T} + \vb{D})^\intercal
\end{equation*}
guaranteed to be \ac{spd}.

Due to restrictions set by the PyTorch implementation the single-precision floating-point format is used for training data $\mathcal{T}$, model parameters and evaluation of $f$.

\section{Results and Discussion}

In this section we show the performance of our proposed preconditioning technique with \ac{ml} based on solving the \ac{ppe} in a \ac{cfd} simulation~\citep{bridson2015fluid} as well as characterizing the hydraulic state in pipe networks with flow continuity and headloss equations in EPANET~2~\citep{rossman2000epanet}.

\subsection*{Poisson's Equation}
The incompressible Navier--Stokes equations describing the motion of viscous fluids read as
\begin{align*}
    \partial_t \vb{u} + (\vb{u}\cdot\gradient) \vb{u} + \frac{1}{\rho} \gradient p &= \vb{F} + \nu \laplacian \vb{u}\,,\\
    \divergence \vb{u} &= 0\,,
\end{align*}
where $\vb{u}$ represents the fluid velocity vector field in $\mathds{R}^2$ or $\mathds{R}^3$, $\rho$ is the density of the fluid, $p$ denotes the scalar pressure field, $\vb{F}$ is the sum of external terms, and $\nu$ is the kinematic viscosity. To this day, existence and smoothness of Navier--Stokes solutions remain an open problem. Thus, various numerical schemes have been developed in the past in order to find satisfactory approximate solutions.

Typically, after some mathematical manipulations, as described in, e.\,g.,~\citep{chorin1967}, enforcing incompressibility of the fluid leads to solving the so-called \ac{ppe} given by
\begin{equation}\label{eq:ppe}
    \laplacian p = \frac{\rho}{\Delta t} \divergence \vb{u}^\ast\,,
\end{equation}
where $\vb{u}^\ast$ is an intermediate solution.

Approximating the Laplacian~$\laplacian$ in~\eqref{eq:ppe} with a finite-difference method in a Eulerian \ac{cfd} simulation results in a large linear system
\begin{equation}\label{eq:discrete}
    \vb{L} \vb{p} = \vb{d}
\end{equation}
equal to~\eqref{eq:linSystem}. The matrix $\vb{L}$ is sparse \ac{spd}, and $\vb{d}$ corresponds to the right-hand side in~\eqref{eq:ppe}. Note that, after a slight modification of the \ac{cnn} $f$, our approach is also suited for particle-based \ac{cfd} methods like \ac{sph} originally developed by~\citet{gingold1977smoothed}. Two types of analytical preconditioners have already been applied to an incompressible \ac{sph} setup on \ac{gpu}s by \citet{chow2018incompressible}.

Solving~\eqref{eq:discrete} and projecting $\vb{u}^\ast$ onto the divergence-free subspace is generally the most expensive and time-consuming part of a fluid simulation~\citep{tompson2016accelerating}. In the following we state the results of increasing \ac{pcg} performance for solving this large system of linear equations by applying our learned preconditioner $\vb{M}^{-1}$ provided by the \ac{cnn} $f$ described in the previous section. In contrast to typical applications of \ac{ml} where total system performance data is analyzed for learning purposes, we are aiming to lessen the computational burden of the numerical solution here.

The \ac{cnn} is trained on \num{800} and validated on \num{200} occupancy grids of size $\num{32}\times \num{32}$ resulting in a $\num{1024}\times \num{1024}$ linear system~\eqref{eq:discrete}. These are generated by intersecting an arbitrarily oriented 2D plane with two mutually exclusive sets of 3D models. We assume homogeneous Dirichlet boundary conditions in all our simulations generated with Mantaflow\footnote{\url{http://mantaflow.com/}}. The geometry is given and fixed, thus our trained \ac{cnn} needs to be evaluated only once for each \ac{cfd} simulation which is achieved in constant time.

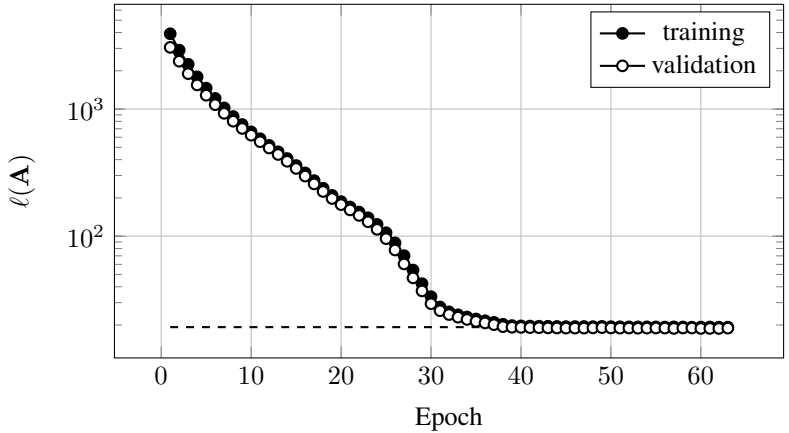
\begin{figure*}[htb]
    \centering
    \usetikzlibrary{plotmarks}
    \begin{tikzpicture}
        \begin{semilogyaxis}[
            grid,
            width=.75\textwidth,
            height=.45\textwidth,
            xlabel={Epoch},
            ylabel={$\ell(\vb{A})$},
            ]
            \addplot[thick,mark=*] table [col sep=comma,header=true] {./ppe_loss.csv};
            \addlegendentry{training};

            \addplot[thick,mark=*,mark options={fill=white}] table [col sep=comma,header=true] {./val_loss.csv};
            \addlegendentry{validation};

            \addplot[thick,dashed,mark=none,domain=1:63] {19.25+0*x};
        \end{semilogyaxis}
    \end{tikzpicture}
    \caption{Semi-log plot of training and validation loss on \num{800} and \num{200} samples, respectively. The model converges to the empirical lower bound of about \num{19.25} (dashed line) which could be observed in several training runs. Note that the validation loss shows no signs of overfitting indicating good generalization properties of the model.}\label{fig:ppe_loss}
\end{figure*}

\begin{figure*}[htb]
    \centering
    \usetikzlibrary{plotmarks}
    \begin{tikzpicture}
        \begin{semilogyaxis}[
            grid,
            width=.75\textwidth,
            height=.45\textwidth,
            xlabel={Iteration},
            ylabel={Residual},
            ]
            \addplot[thick,mark=*] table [col sep=comma,header=true,x index=0,y index=1] {./res_vanilla.csv};
            \addlegendentry{vanilla};

            \addplot[thick,mark=*,mark options={fill=white}] table [col sep=comma,header=true,x index=0,y index=1] {./res_jacobi.csv};
            \addlegendentry{Jacobi};

            \addplot[thick,mark=square*] table [col sep=comma,header=true,x index=0,y index=1] {./res_ic_0_.csv};
            \addlegendentry{IC(0)};

            \addplot[thick,mark=square*,mark options={fill=white}] table [col sep=comma,header=true,x index=0,y index=1] {./res_amg.csv};
            \addlegendentry{AMG};

            \addplot[thick,mark=triangle*] table [col sep=comma,header=true,x index=0,y index=1] {./res_learned.csv};
            \addlegendentry{learned};

            \addplot[thick,dashed,domain=09:26] {10^(-x/2.45-.5)};
            \addlegendentry{slope \num{2.45}};
        \end{semilogyaxis}

    \end{tikzpicture}
    \caption{We consider a non-cherry-picked \num{4096}-dimensional discrete Poisson problem and compare the residuals in a semi-log plot. Vanilla \ac{cg} and Jacobi \ac{pcg} can not be distinguished since they converge at the same speed, incomplete Cholesky with zero fill-in (IC(0)) is slightly better. Our preconditioner (learned) is \SI{2.172}{\percent} dense, converges with order \num{2.45}, and outperforms the \ac{amg} approach in terms of runtime, cf. Table~\ref{tab:summary_big}.}\label{fig:residuals}
\end{figure*}
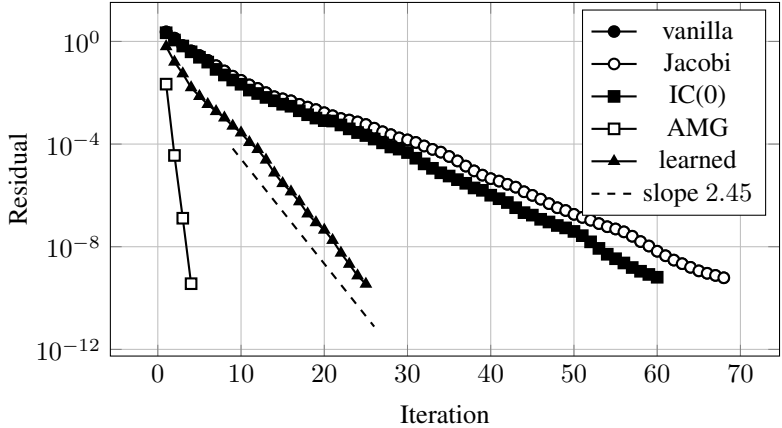

Training and validation of this model with a total of \num{1180934} internal parameters took \SI{9.5}{\hour} on an NVIDIA~Titan~X \ac{gpu} for \num{64} epochs, although it already converged at \num{40} epochs. The overall optimization progress is plotted in Figure~\ref{fig:ppe_loss}.
In Figure~\ref{fig:residuals} we compare the residuals of analytical preconditioning approaches, such as Jacobi, incomplete Cholesky with zero-fill, and \ac{amg}, with our learned preconditioner. Clearly, the \ac{cnn} is quite capable of designing $\vb{M}^{-1}$ in such a way that \ac{pcg} converges in less than half as much iterations.
In Table~\ref{tab:summary_small} the quantitative results for \ac{pcg} averaged over \num{200} samples not used for training can be observed. The time needed for constructing the preconditioners is negligible and has thus been ignored.
\begin{table*}[htb]
    \centering
    \begin{tabular}{*5c}
        \toprule
        method & \text{time} [\si{\ms}] & iter & $\kappa$ & density\\\midrule
        vanilla & \num{9.829} & \num{54.64} & \num{139.367} & --\\
        jacobi & \num{9.695} & \num{54.64} & \num{139.367} & {\bfseries\SI[detect-weight]{0.134}{\percent}}\\
        ic(0) & \num{8.773} & \num{47.41} & \num{114.275} & \SI{2.864}{\percent}\\
        {\bfseries amg} & {\bfseries \num[detect-weight]{3.082}} & {\bfseries\num[detect-weight]{4.02}} & {\bfseries \num[detect-weight]{1.261}} & \SI{98.686}{\percent}\\
        learned & \num{4.827} & \num{21.07} & \num{18.904} & \SI{5.749}{\percent}\\
        \toprule
    \end{tabular}
    \caption{Comparison of three standard preconditioning approaches with our learned \ac{ml} model. The underlying problem is \num{1024}-dimensional, all results are averaged over \num{200} samples. Since the size is so small all other preconditioners fail to beat the \ac{amg} approach in terms of runtime and iterations.}\label{tab:summary_small}
\end{table*}
Furthermore, it is important to note that once the \ac{cnn} is evaluated we can reuse $\vb{M}^{-1}$ in each time step whenever we need to solve~\eqref{eq:discrete} resulting in a linear increase of computation time saved as the simulation progresses.

In order to investigate the generalization capabilities of our model with respect to matrix dimensions we tested it on a \num{4096}-dimensional problem and averaged over \num{80} samples. Even though it was originally trained on a small fluid domain it scales to bigger ones quite well, as can be seen in Table~\ref{tab:summary_big}.
\begin{table*}[htb]
    \centering
    \begin{tabular}{*5c}
        \toprule
        method & \text{time} [\si{\ms}] & iter & $\kappa$ & density\\\midrule
        vanilla & \num{22.70} & \num{98.15} & \num{472.727} & --\\
        jacobi & \num{27.02} & \num{95.15} & \num{472.727} & {\bfseries\SI[detect-weight]{0.032}{\percent}}\\
        ic(0) & \num{56.17} & \num{83.24} & \num{398.466} & \SI{3.671}{\percent}\\
        amg & \num{52.02} & {\bfseries\num[detect-weight]{4.72}} & {\bfseries \num[detect-weight]{1.539}} & \SI{99.471}{\percent}\\
        {\bfseries learned} & {\bfseries \num[detect-weight]{14.25}} & \num{35.86} & \num{61.494} & \SI{1.413}{\percent}\\
        \toprule
    \end{tabular}
    \caption{Results for the same model that was used to generate Table~\ref{tab:summary_small}, averaged over \num{80} samples. The \num{4096}-dimensional problems here are four times as large as the ones in the training data. The high sparsity of our learned preconditioner induces considerable improvement in terms of runtime compared to all the other techniques.}\label{tab:summary_big}
\end{table*}
The higher sparsity in comparison with an \ac{amg} approach causes each \ac{pcg} iteration to be computationally more efficient and consequently faster.

\subsection*{EPANET~2}
The open-source toolkit EPANET~2~\citep{rossman2000epanet} developed by the U.\,S. Environmental Protection Agency is a software package for modeling pressure and flow conditions in a drinking water network. Assuming the network consists of $n$ junction nodes a linear system~\eqref{eq:linSystem} that reads as
\begin{equation*}\label{eq:epanet_linsys}
    \vb{A}\vb{h} = \vb{f}\,,
\end{equation*}
where $\vb{A} \in \R^{n \times n}$ has to be solved for unknown nodal heads $\vb{h}$. It can be shown that $\vb{A}$ is a real \ac{spd} matrix, thus suited for the \ac{pcg} algorithm. \citet{zecchin2012steady} found that \ac{amg} preconditioning outperforms the EPANET~2 solver if the networks are very large. \cite{burger2015quest} accelerate the hydraulic simulations by utilizing a multicore capable solver but could not achieve satisfying results for networks with a real-world character.

We split a data set containing artificial water networks with a mimimum of \num{1024} and up to \num{3971} junctions into \num{442} and \num{110} samples for training and testing, respectively. A revised version of the model employed for the \ac{ppe} was used. Due to increased scattering of non-zero entries in~$\vb{A}$ the learned preconditioner is much more dense than it would be if $\vb{A}$ resembled a band matrix resulting in slower evaluation of each \ac{pcg} iteration. The overall results were inconclusive with even analytical preconditioning approaches failing in some cases. Special care is needed, e.\,g., reordering the input matrix to cluster non-zeros or adjusting the hyperparameters of the \ac{cnn} model, which would be beyond the scope of this paper.

\section{Conclusions}

The rise of \ac{ml} has mainly revolutionized image processing and data analysis techniques. We proposed applying these methods, namely a \ac{cnn}, to solving large sparse systems of linear equations. However, the approximation error usually being introduced by \ac{ml} approaches is undesirable in real-world engineering applications. By relying on the well-known iterative \ac{pcg} algorithm we were able to utilize a \ac{cnn} to design preconditioners, while still guaranteeing numerical accuracy of the obtained solutions. This novel approach has been found to work excellent for a variety of problems and in some instances even results in speed-up factors of about \numrange{2}{3} for a single execution. We have been able to demonstrate the feasibility of our approach for the case study of \ac{cfd}.

Despite the initial success, there still exist areas for continued development. Finding a replacement for the condition number computed via the costly \ac{svd} would accelerate the training process. We also intend to investigate learned preconditioning for more general types of equations not necessarily resulting in linear systems with \ac{spd} matrices.

\subsubsection*{Acknowledgments}

We thank Gregor Burger for making his EPANET~2 code available. This research is part of the SPHAUL project 850738 which is funded by the Austrian Research Promotion Agency (FFG). We gratefully acknowledge the support of NVIDIA Corporation with the donation of two Titan~X \ac{gpu}s used for this publication.

\bibliographystyle{apalike}
{\footnotesize \bibliography{references}}

\end{document}